\documentclass[10pt,twocolumn,letterpaper]{article}

\usepackage[pagenumbers]{cvpr} 

\definecolor{cvprblue}{rgb}{0.21,0.49,0.74}
\usepackage[pagebackref,breaklinks,colorlinks,allcolors=cvprblue]{hyperref}
\usepackage{indentfirst}

\def\paperID{1991} 
\def\confName{CVPR}
\def\confYear{2025}

\title{Identity-Preserving Text-to-Image Generation via Dual-Level Feature Decoupling and Expert-Guided Fusion}

\author{
  Kewen Chen\textsuperscript{1} \quad
  Xiaobin Hu\textsuperscript{2} \quad
  Wenqi Ren\textsuperscript{1}\thanks{Corresponding author.} \\
  \textsuperscript{1}School of Cyber Science and Technology, Shenzhen Campus of Sun Yat-sen University \\
  \textsuperscript{2}Technische Universität München \\
  {\tt\small chenkw23@mail2.sysu.edu.cn, xiaobin.hu@tum.de, renwq3@mail.sysu.edu.cn}
}

\begin{document}
\maketitle
\begin{abstract}
\indent
Recent advances in large-scale text-to-image generation models have led to a surge in subject-driven text-to-image generation, which aims to produce customized images that align with textual descriptions while preserving the identity of specific subjects. Despite significant progress, current methods struggle to disentangle identity-relevant information from identity-irrelevant details in the input images, resulting in overfitting or failure to maintain subject identity. In this work, we propose a novel framework that improves the separation of identity-related and identity-unrelated features and introduces an innovative feature fusion mechanism to improve the quality and text alignment of generated images. Our framework consists of two key components: an Implicit-Explicit foreground-background Decoupling Module (IEDM) and a Feature Fusion Module (FFM) based on a Mixture of Experts (MoE). IEDM combines learnable adapters for implicit decoupling at the feature level with inpainting techniques for explicit foreground-background separation at the image level. FFM dynamically integrates identity-irrelevant features with identity-related features, enabling refined feature representations even in cases of incomplete decoupling. In addition, we introduce three complementary loss functions to guide the decoupling process. Extensive experiments demonstrate the effectiveness of our proposed method in enhancing image generation quality, improving flexibility in scene adaptation, and increasing the diversity of generated outputs across various textual descriptions.
\end{abstract}    
\section{Introduction}
\label{sec:introduction}

\begin{figure*}[thbp] \centering
    \includegraphics[width=\textwidth,height=0.27\textwidth]{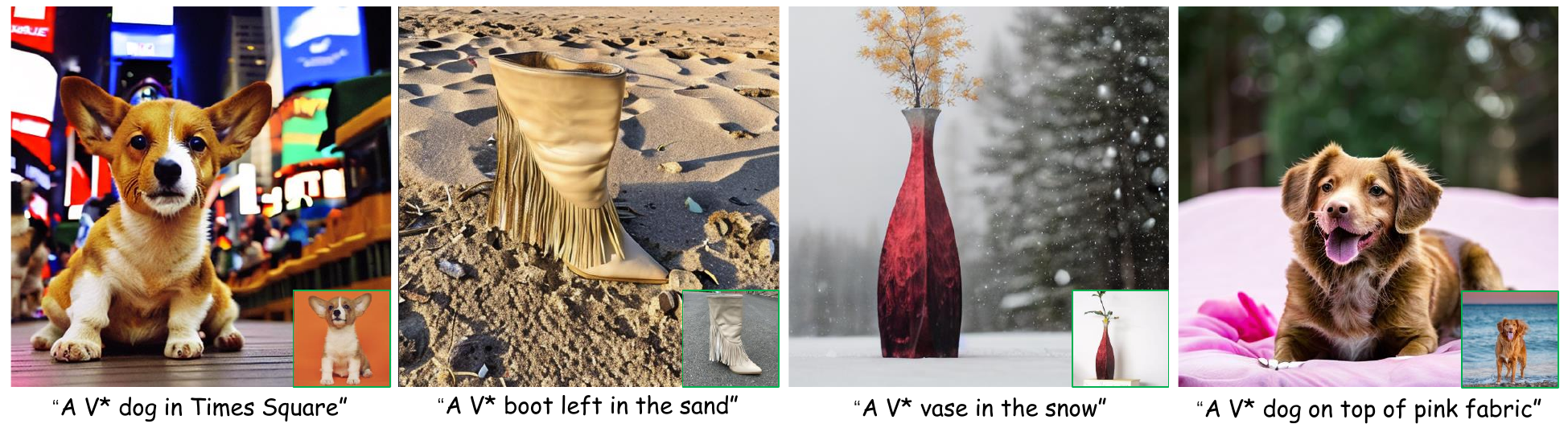}
    \caption{\textbf{Example images generated by our proposed method.} Our approach produces high-quality images that maintain identity consistency while aligning with the input text prompts.} \label{fig:intro}
\end{figure*}

\indent
Recently, large-scale text-to-image generation models have achieved remarkable progress \cite{nichol2021glide,peebles2023scalable,dhariwal2021diffusion_models_beat_gans,saharia2022photorealistic,podell2023sdxl,rombach2022stable_diffusion}. Taking advantage of the exceptional image generation capabilities of these models, subject-driven text-to-image generation - also known as customized image generation - has garnered widespread attention \cite{gal2022textual_inversion,ruiz2023dreambooth,zhu2024multibooth,kumari2025generating,kumari2023custom_diffusion,chen2023disenbooth,pang2024attndreambooth,han2023svdiff,kong2024omg,shah2024ziplora,park2024textboost}. This task aims to fine-tune a pre-trained text-to-image generation model, \eg, Stable Diffusion \cite{rombach2022stable_diffusion}, using a few reference images of a specific subject. The goal is to enable the model to generate images that not only align with a given textual description but also retain the unique visual characteristics of the specified subject.

Despite substantial advancements in subject-driven text-to-image generation techniques, existing methods \cite{ruiz2023dreambooth,kumari2023custom_diffusion,pang2024attndreambooth, park2024textboost} often struggle to effectively separate identity-relevant information from identity-irrelevant details within the input images. This limitation leads to generated images that either disregard textual prompts and overfit to the input images or fail to preserve the subject's identity. Although some methods \cite{park2024textboost,chen2023disenbooth,cai2024decoupled} have attempted to disentangle identity-related and identity-irrelevant information, such as TextBoost~\cite{park2024textboost}, which employs image augmentation strategies and introduces augmentation tokens associated with specific image augmentation types,  they do not address the foreground and background, resulting in incomplete disentanglement and susceptibility to "augmentation leaking" phenomena. DisenBooth~\cite{chen2023disenbooth} introduces an Identity-Irrelevant Branch that uses a learnable mask to separate identity-related information from image features, but this implicit disentanglement may not effectively learn a robust representation of identity-irrelevant features. Moreover, while DisenBooth~\cite{chen2023disenbooth} considers the decoupling of the foreground and background, it simply combines the decoupled features without a strategy for better integration, leading to combined features that do not correspond well to the original images, thereby adversely affecting the generation process.

To address these challenges, we propose a novel framework that enhances the disentanglement of identity-related and identity-irrelevant features and introduces an innovative feature fusion mechanism to improve the quality and text alignment of generated images. Our framework consists of two key components: a hybrid Implicit-Explicit foreground-background Decoupling Module (IEDM) and a Feature Fusion Module (FFM) based on a Mixture of Experts (MoE) model. Specifically, the IEDM employs a learnable adapter at the feature level to extract identity-irrelevant features, achieving implicit decoupling. At the image level, it leverages current inpainting techniques \cite{yu2023inpaint_anything} to separate the foreground subject from the background, utilizing the background information to further reinforce the extraction of identity-irrelevant features, achieving explicit decoupling. This dual-level approach enhances the model's ability to capture identity-irrelevant details. Furthermore, to ensure effective separation of identity-relevant and identity-irrelevant information, we propose three complementary loss functions to guide the decoupling process.

The MoE-based FFM then integrates the identity-irrelevant background features with the identity-related foreground features. It allows the model to dynamically adjust its focus on different features, amplifying foreground features under the guidance of multiple experts while compressing background information, thereby optimizing feature representation. Even when identity-related information is present in the background, this module can mitigate the impact on the overall results when the foreground and background are not cleanly decoupled. We conducted extensive experiments to evaluate our method against state-of-the-art baselines, demonstrating its effectiveness in generating high-quality images that align with textual descriptions while preserving the subject's identity.

In summary, our contributions can be outlined as follows:

\begin{itemize}[label=\textbullet]
\item We propose an Implicit-Explicit Foreground-Background Decoupling Module (IEDM) that integrates implicit decoupling at the feature level with explicit decoupling at the image level, ensuring a more thorough separation of foreground and background. In this process, we design three complementary loss functions to guide the decoupling process.
\item We introduce a Feature Fusion Module (FFM) based on a Mixture of Experts (MoE) model, which dynamically integrates identity-irrelevant background features with identity-related foreground features. This fusion approach enables the model to provide refined feature representations, even in cases of incomplete decoupling.
\item Extensive experiments demonstrate the effectiveness of our proposed method in enhancing image generation quality, improving flexibility in scene adaptation, and increasing the diversity of generated outputs across various textual descriptions.
\end{itemize}

\section{Related Work}
\label{sec:relatedwork}

\subsection{Text to Image Generation}

\indent
Generating images from textual descriptions has been a long-standing goal in the field of artificial intelligence, driving the development of various generative models \cite{podell2023sdxl,rombach2022stable_diffusion,saharia2022photorealistic,ramesh2022hierarchical,nichol2021glide, hu2024diffumatting, avrahami2022blended, bar2022text2live, brooks2023instructpix2pix,balaji2022ediff,esser2024scaling}. Early approaches primarily utilized Generative Adversarial Networks (GANs)~\cite{goodfellow2020GAN,kocasari2022stylemc,patashnik2021styleclip,xu2022predict}, which demonstrated the ability to translate text into corresponding images with remarkable success. However, GANs often face challenges in precisely controlling the generated content, leading to issues such as unnatural artifacts and a lack of fine-grained control over the generated images.
The advent of diffusion models~\cite{ho2020ddpm} marked a significant shift in the paradigm of text-to-image generation. These models operate by gradually adding noise into the data and then learning to reverse this process, enabling the generation of images conditioned on textual prompts. Compared to GANs, diffusion models have demonstrated the ability to produce higher fidelity and more diverse outputs, offering a more flexible and controllable generation process~\cite{dhariwal2021diffusion_models_beat_gans}. 
Models like Stable Diffusion~\cite{rombach2022stable_diffusion}, trained on large-scale datasets, have emerged as state-of-the-art techniques for text-to-image generation, opening new avenues for creative expression and practical applications, such as virtual try-on \cite{kim2024stableviton,zeng2024cat,baldrati2023multimodal,morelli2023ladi}, personalized content creation \cite{ruiz2023dreambooth, huang2024consistentid, ma2024subject}, and artistic exploration \cite{layout2024cvpr,layout2024eccv,layout2024iclr}.

\subsection{Subject-Driven Customization}
Subject-driven text-to-image generation \cite{shi2024instantbooth,ruiz2023dreambooth,chen2023disenbooth,gal2022textual_inversion,park2024textboost, cai2024decoupled,kumari2023custom_diffusion,pang2024attndreambooth,yang2024dreammix,ram2025dreamblend,xiong2024groundingbooth,shi2024relationbooth} employs personalized fine-tuning techniques to associate specific visual identities with unique text tokens. Given a small set of reference images, methods like Textual Inversion~\cite{gal2022textual_inversion} and DreamBooth~\cite{ruiz2023dreambooth} introduce identity tokens (\eg, "V*") that guide the model to generate personalized images consistent with textual prompts while retaining identity fidelity. For instance, Textual Inversion~\cite{gal2022textual_inversion} learns an embedding for the identity token by optimizing it on user-provided images, whereas DreamBooth~\cite{ruiz2023dreambooth} fine-tunes the entire model to enhance personalization. AttnDreamBooth~\cite{pang2024attndreambooth} combines Textual Inversion and DreamBooth, using a multi-stage training strategy and introducing a cross-attention map regularization term to achieve personalization. Recent methods \cite{park2024textboost,chen2023disenbooth} attempt to disentangle identity-relevant and identity-irrelevant information to improve identity preservation. DisenBooth~\cite{chen2023disenbooth} introduces an identity-irrelevant branch that uses a learnable mask to separate subject identity from other image features. TextBoost~\cite{park2024textboost} focuses on fine-tuning only the text encoder. It employs data augmentation strategies and introduces augmentation tokens associated with specific types of image transformations to disentangle identity-relevant and identity-irrelevant features.

Despite these advancements, current methods still face challenges in fully decoupling identity from context, and there remains room for improvement in the integration of disentangled features. Other tuning-free methods~\cite{wang2024instantid,chen2024anydoor,he2025anystory,ye2023ip_adapter,kong2024anymaker,li2023blip,zhang2024ssr,huang2024consistentid,ma2024subject,song2024moma,wang2024ms}, such as IP-Adapter~\cite{ye2023ip_adapter} and AnyMaker~\cite{kong2024anymaker}, can achieve inference through a single forward propagation; however, they require extensive data and significant computational resources for robust training. Our approach introduces a more refined disentanglement strategy with enhanced extraction of identity-irrelevant features and flexible feature fusion.

\subsection{Feature Fusion and Mixture of Experts (MoE)}
\indent

The effective fusion of disparate features is crucial in enhancing the performance of text-to-image generation models, particularly when it comes to subject-driven customization. Feature fusion aims to combine complementary information from different sources to improve the model's ability to generate images that are both contextually relevant and visually coherent. One of the most promising approaches in this domain is the Mixture of Experts (MoE) model \cite{moe2013,jacobs1991adaptive_moe,shen2023scaling_vl_moe,moe2017,moe2020,moe2021,moe2022switchtransformer,cai2024survey_moe,moe2023,fastermoe2022}, which has been increasingly adopted to address the challenges of integrating diverse features in a computationally efficient manner. 
MoE models, initially introduced by Jacobs \etal.~\cite{jacobs1991adaptive_moe}, are a type of ensemble learning algorithm where each "expert" is a neural network specialized in a particular subset of the data. The key innovation of MoE lies in its gating mechanism, which dynamically routes inputs to the most suitable expert based on their content. This not only allows for efficient computation by activating only relevant experts but also enables the model to leverage the strengths of diverse feature representations.
Shen \etal.~\cite{shen2023scaling_vl_moe} utilizes MoE to combine textual embeddings with image features, allowing the model to focus on different regions of the input image based on the textual prompt.

Our Feature Fusion Module (FFM) builds upon these insights by adopting a MoE framework to integrate identity-irrelevant background features with identity-related foreground features. Unlike previous works that simply concatenate or average features, our FFM leverages the MoE gating mechanism to dynamically weight the contributions of different features, allowing the model to adaptively focus on the most relevant information for generating the final image. This approach not only optimizes the feature representation but also mitigates the impact of incomplete foreground-background decoupling, leading to improved image generation quality and textual alignment.

\section{Method}
\label{sec:method}

In this work, we propose a novel framework that enhances the separation of identity-related and identity-irrelevant features and incorporates a feature fusion mechanism to refine the extracted features. The framework consists of two key components: the Implicit-Explicit foreground-background Decoupling Module (IEDM) and the MoE-based Feature Fusion Module (FFM). Below, we detail the working of each component.

\subsection{Overview}

\begin{figure*}
  \centering
  \includegraphics[width=\linewidth]{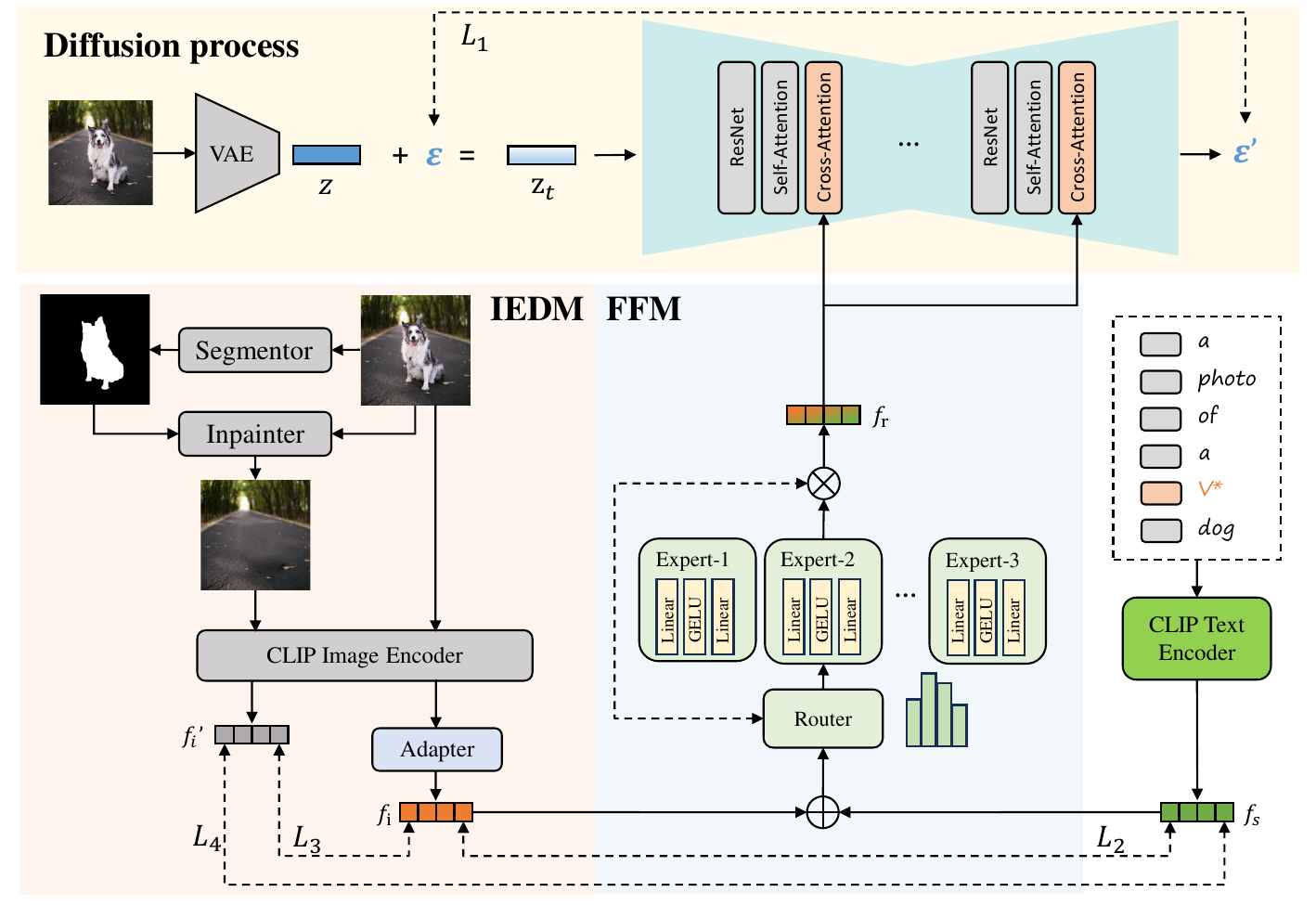}
    \caption{\textbf{Overview of our proposed method.} The framework consists of the Implicit-Explicit foreground-background Decoupling Module (IEDM) for separating identity-related and identity-irrelevant features, and the Mixture of Experts (MoE)-based Feature Fusion Module (FFM) for refining the combined feature representations. The process begins with a text prompt that generates identity-related features, followed by dual-level decoupling of the input image to extract identity-irrelevant background features. These features are then integrated through the MoE-based FFM, and the refined feature representations are used as conditioning input for the U-Net denoising process to produce high-quality images.}
    \label{fig:overview}
\end{figure*}

The overall framework of our proposed method is illustrated in Figure~\ref{fig:overview}. Given a prompt $P$ containing a specific identifier, \eg, "a photo of a V* dog," where "V*" designates the subject we aim to bind, the CLIP text encoder \cite{clip} processes $P$ to generate text features $f_s$. Since the prompt $P$ is shared across all input images $\{x_i\}$, $f_s$ captures identity-related foreground information and serves as a representation of identity-relevant features~\cite{chen2023disenbooth}. The input image $x_i$ undergoes dual-level decoupling in the IEDM, resulting in identity-irrelevant background features $f_i$. Subsequently, $f_i$ and $f_s$ are combined and processed through the FFM to obtain a refined feature representation $f_{r}$. This refined feature $f_{r}$ then serves as a conditioning input to guide the U-Net denoising process.

\subsection{Implicit-Explicit foreground-background Decoupling Module}

The IEDM takes the input image $x_i$ and enhances the separation of identity-related and identity-irrelevant features through a dual-level decoupling process, yielding identity-irrelevant background features $f_i$.

\noindent
{\bf Implicit Decoupling:} As shown in Figure~\ref{fig:overview}, this process begins with a pretrained CLIP image encoder \cite{clip} $E_I$ that extracts feature representations $f^{(p)}_i = E_I(x_i)$ from an input image $x_i$. At this stage, $f^{(p)}_i$ contains both identity-relevant and identity-irrelevant information. Next, an adapter is employed to implicitly extract identity-irrelevant features. This adapter comprises a learnable mask, with values between $(0,1)$ and dimensions matching the feature representation, along with several linear layers equipped with skip connections. The adapter selectively filters out identity-relevant information from $f^{(p)}_i$, focusing on capturing features that are not directly related to the subject's identity. The adapter ultimately outputs identity-irrelevant background features $f_i$, achieving implicit decoupling at the feature level. The process is formally represented as follows:
\begin{equation}
f_i = \text{Adapter}(E_I(x_i)), \quad i=1,2,\dots,n
\end{equation}
where $E_I$ denotes the CLIP Image Encoder, $x_i$ is the input image, and $n$ is the number of images in the small image set.

Furthermore, since $f_i$ and $f_s$ represent identity-irrelevant and identity-related features, respectively, their similarity should be minimized. To ensure that implicit decoupling accurately captures identity-irrelevant features, we employ a contrastive loss to guide this process:
\begin{equation}
L_2 = \sum_{i=1}^{n} \text{cos}(f_i, f_s)
\end{equation}
where $\text{cos}$ denotes the cosine similarity loss.

\noindent
{\bf Explicit Decoupling:} In this step, we first use the segmentation model~\cite{zhang2024efficientvit_sam} to derive the mask $M_i$ from the input image $x_i$. Subsequently, we feed $x_i$ and $M_i$ into the inpainting module, utilizing the state-of-the-art inpainting model~\cite{suvorov2022LaMa_inpainting} to explicitly separate the foreground subject from the background, yielding an image $x_i'$ that contains only the background. We then encode the inpainted image $x_i'$, which retains only the background, using the CLIP image encoder to obtain the background features $f_i'$, achieving explicit decoupling at the image level. This process can be formalized as follows:
\begin{equation}
f_i' = E_I(\text{Inpaint}(x_i, M_i)), \quad i=1,2,\dots,n
\end{equation}
where $\text{Inpaint}$ denotes the inpainting model, and $M_i$ represents the mask extracted from the input image $x_i$.

Subsequently, we utilize $f_i'$ to enhance the extraction of identity-irrelevant and identity-related features. To achieve this, we introduce two additional contrastive loss terms:
\begin{equation}
L_3 = -\sum_{i=1}^{n} \text{cos}(f_i, f_i')
\end{equation}

Optimizing this loss term encourages $f_i$ to more accurately capture identity-irrelevant features. Similarly, we construct a loss between $f_i'$ and $f_s$ to ensure that $f_s$ more precisely captures identity-related features:
\begin{equation}
L_4 = \sum_{i=1}^{n} \text{cos}(f_s, f_i')
\end{equation}

\subsection{Feature Fusion Module}
Following the decoupling process, we introduce the Feature Fusion Module (FFM) to integrate the identity-irrelevant background features $f_i$ with the identity-related foreground features $f_s$. This module is built upon a Mixture of Experts (MoE) model \cite{moe2022switchtransformer}, which allows for dynamic adjustment of focus on different features, enabling the model to amplify significant features while compressing less relevant ones. The decoupled foreground and background features are first combined and then fed into the FFM module. 

The FFM consists of a gating module $R$ and a set of expert networks $\{Expert_i\}_{i=1}^k$, each specializing in different aspects of feature processing. The outputs of the expert networks are weighted by the gating module $R$, which learns to balance each expert’s contribution based on the input features. This produces a refined set of features representing a balanced integration of foreground and background information, which is then used to generate the final image.
Mathematically, the feature fusion process is expressed as:
\begin{equation}
f_{r} = \sum_{i=1}^k R(f_{com})_i \cdot Expert_i(f_{com}),
\end{equation}
where $f_{com} = f_s + f_i'$ represents the combined feature input, with $Expert_i(\cdot)$ denoting the $i$-th expert network in the FFM module. The gating function $R(f_{com})_i$ provides a weight that modulates the contribution of each expert network based on the input $f_{com}$. Here, $R$ satisfies the constraint $\sum_{i=1}^k R(f_{com})_i = 1$.

This weighted summation of expert outputs results in $f_{r}$, a refined feature representation that effectively balances foreground and background information.

\subsection{Training Strategy}

\noindent
{\bf Training objective:} During training, we use $f_r$ as the conditioning input to the U-Net to reconstruct the image:
\begin{equation}
L_1 = \left\| \epsilon - \epsilon_\theta(z_t, t, f_r) \right\|^2_2
\end{equation}
where $\epsilon$ is random Gaussian noise, $\epsilon_\theta$ denotes the denoising network, $t$ is the sampled time step, and $z_t$ represents the noisy latent of the image $x_i$. 

The total training objective is formulated as:
\begin{equation}
L = \lambda_1 L_1 + \lambda_2 L_2 + \lambda_3 L_3 + \lambda_4 L_4
\end{equation}
where $\lambda_1$, $\lambda_2$, $\lambda_3$, and $\lambda_4$ are the weights for $L_1$, $L_2$, $L_3$, and $L_4$, respectively.

\noindent
{\bf Fine-tuning parameters:} Based on the findings of~\cite{park2024textboost, kumari2023custom_diffusion}, the parameters of the UNet cross-attention layers and the text encoder undergo the most significant changes during fine-tuning. Therefore, we apply LoRA~\cite{hu2021LoRa} to the cross-attention layers of the UNet and the text encoder to achieve improved performance and efficient fine-tuning. The complete set of fine-tuning parameters includes the text embedding for $V*$, the adapter module in the IEDM, the FFM module, and the LoRA parameters within both the UNet and the text encoder.

\noindent
{\bf Inference:} During the inference phase, given a custom prompt $P'$ that encompasses identity information and and describes various background contents, we use only the text features $E_T(P')$ encoded by the text encoder as conditional input to generate high-quality images aligned with the text prompts.

\section{Experiments}
\label{sec:experiments}

\begin{figure*}[tb!p] \centering
    \makebox[0.13\textwidth]{\footnotesize Input Image}
    \makebox[0.13\textwidth]{\footnotesize Textual Inversion}
    \makebox[0.13\textwidth]{\footnotesize DreamBooth}
    \makebox[0.13\textwidth]{\footnotesize AttnDreamBooth}
    \makebox[0.13\textwidth]{\footnotesize DisenBooth}
    \makebox[0.13\textwidth]{\footnotesize TextBoost}
    \makebox[0.13\textwidth]{\footnotesize Ours}
    \\
    \includegraphics[width=0.13\textwidth]{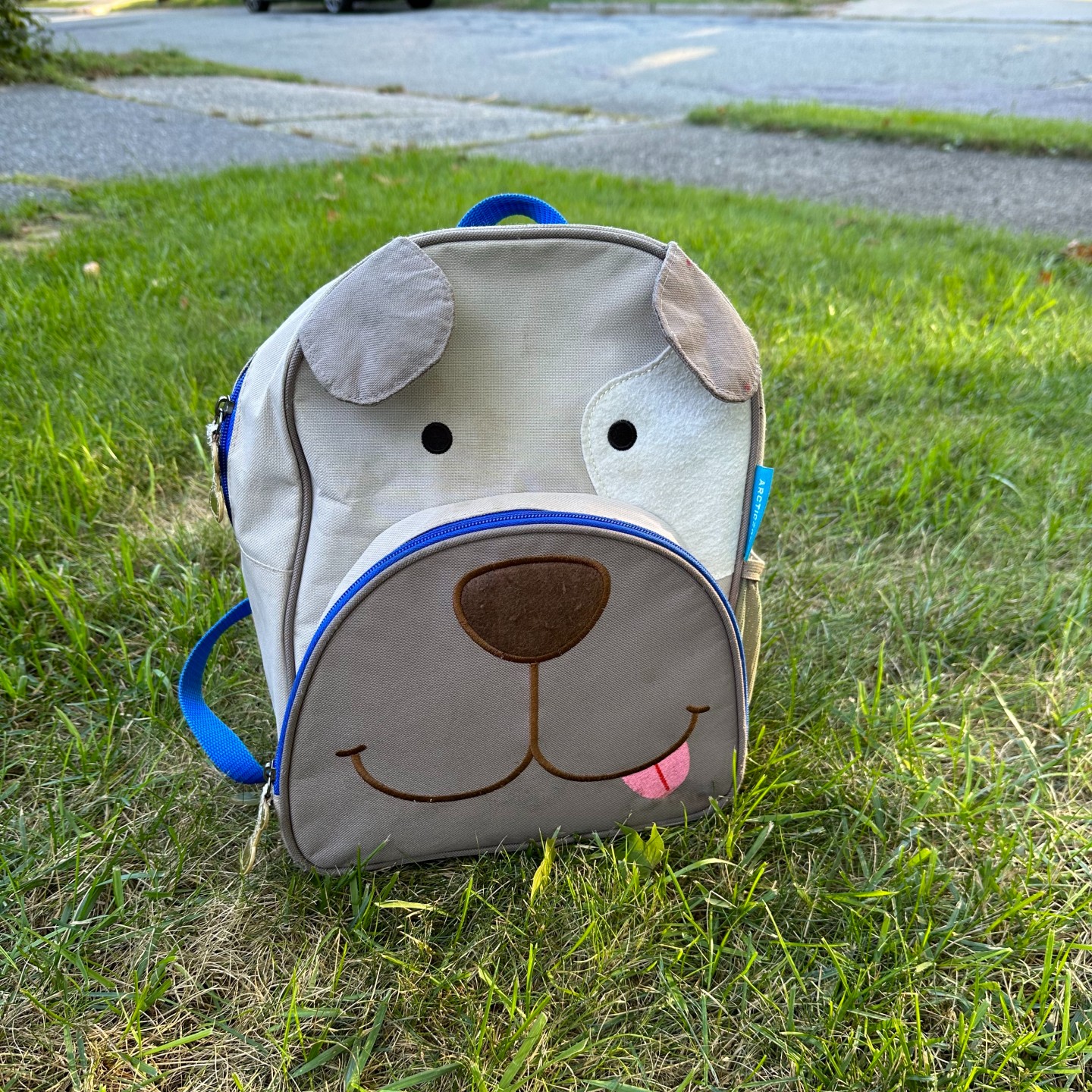}
    \includegraphics[width=0.13\textwidth]{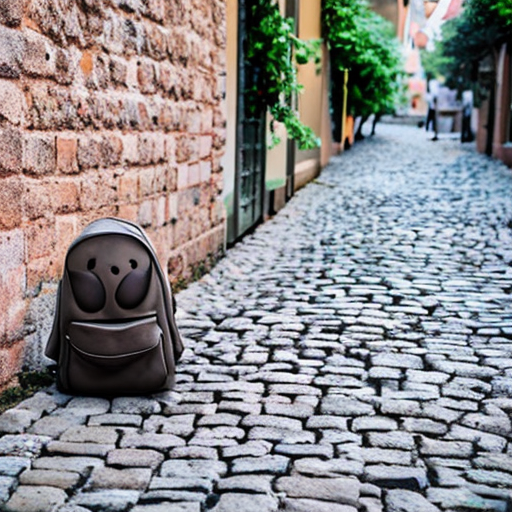}
    \includegraphics[width=0.13\textwidth]{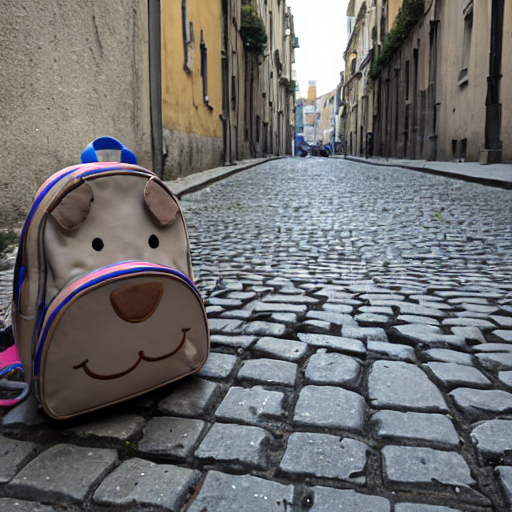}
    \includegraphics[width=0.13\textwidth]{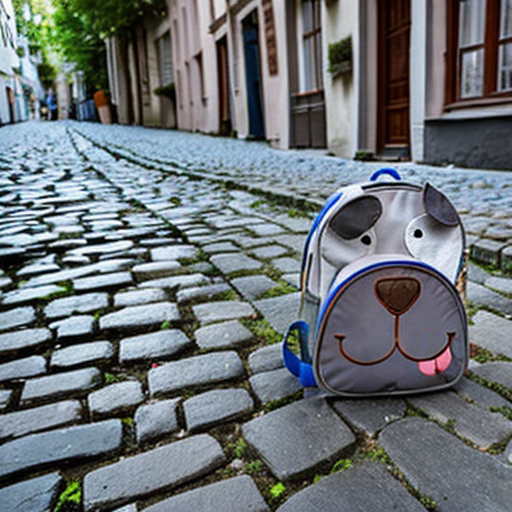}
    \includegraphics[width=0.13\textwidth]{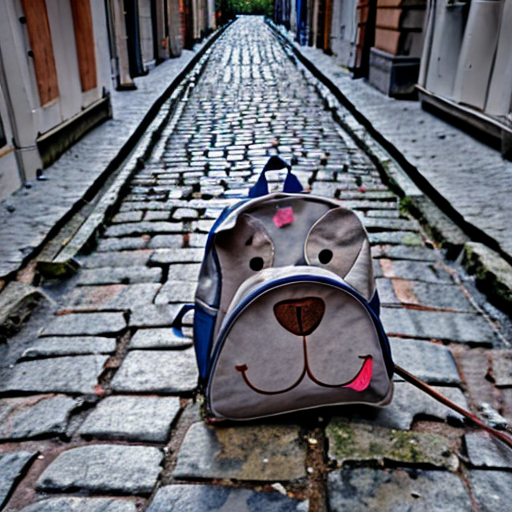}
    \includegraphics[width=0.13\textwidth]{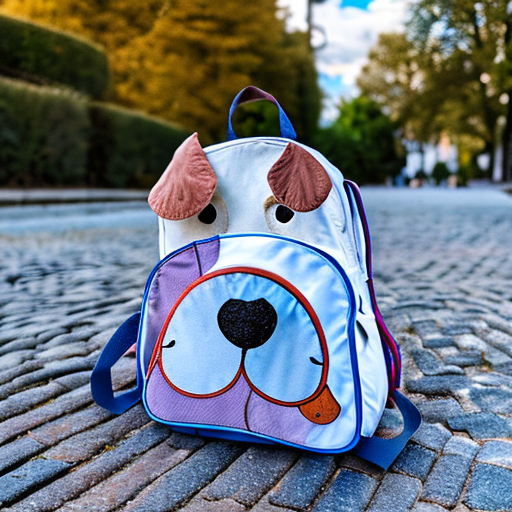}
    \includegraphics[width=0.13\textwidth]{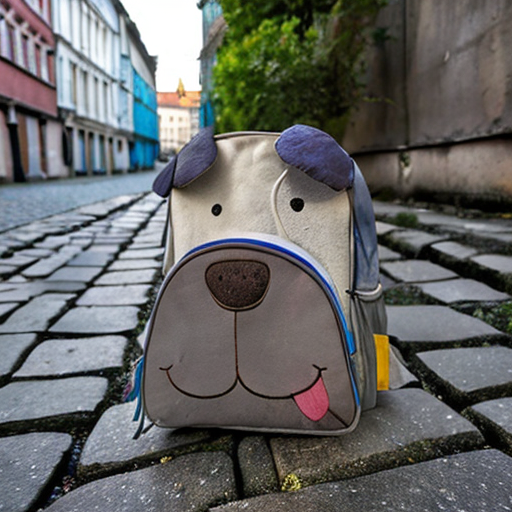}
    \\
    \makebox[\textwidth]{\sffamily "A V* backpack on a cobblestone street."}
    \\
    \includegraphics[width=0.13\textwidth]{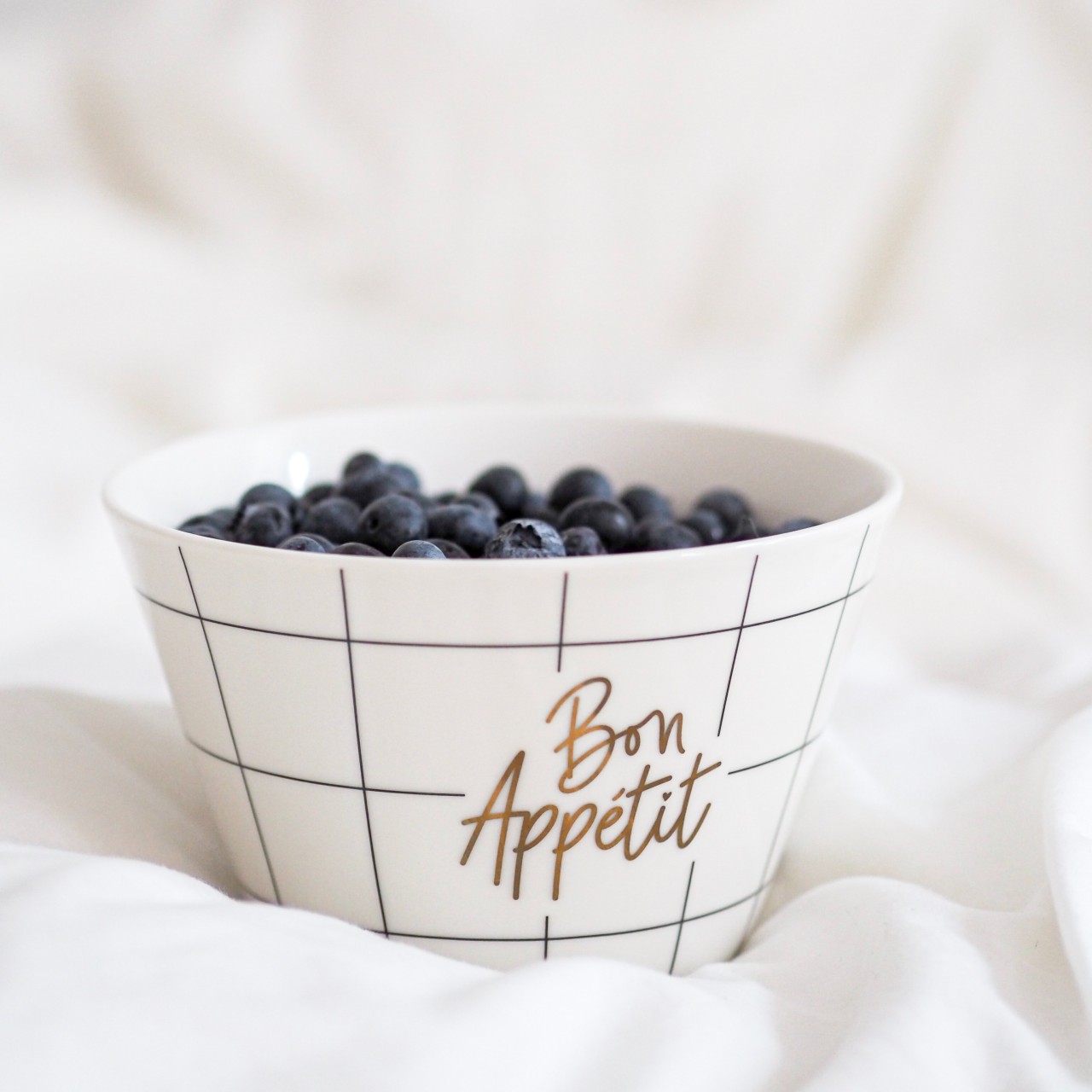}
    \includegraphics[width=0.13\textwidth]{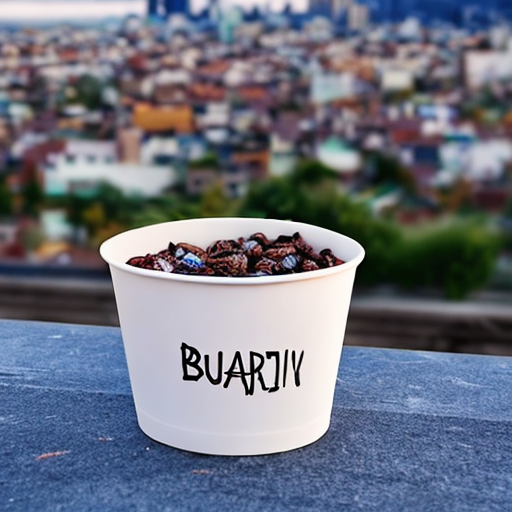}
    \includegraphics[width=0.13\textwidth]{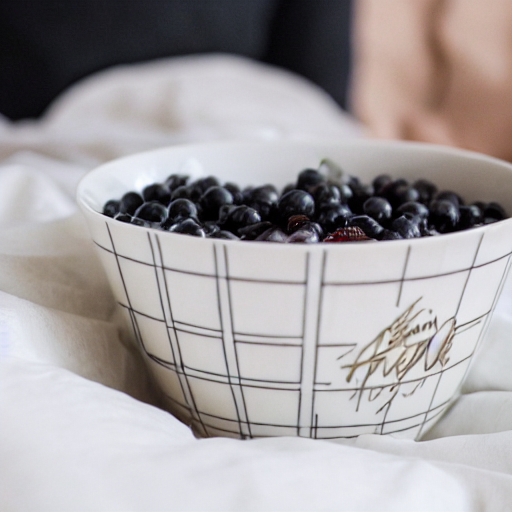}
    \includegraphics[width=0.13\textwidth]{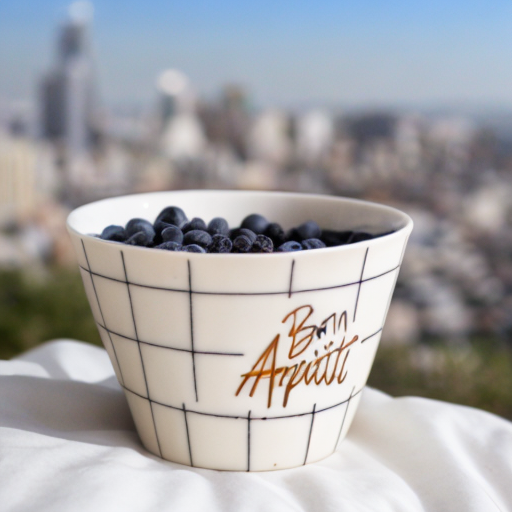}
    \includegraphics[width=0.13\textwidth]{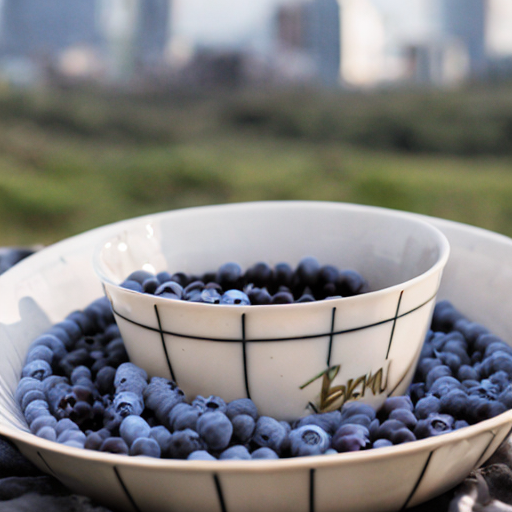}
    \includegraphics[width=0.13\textwidth]{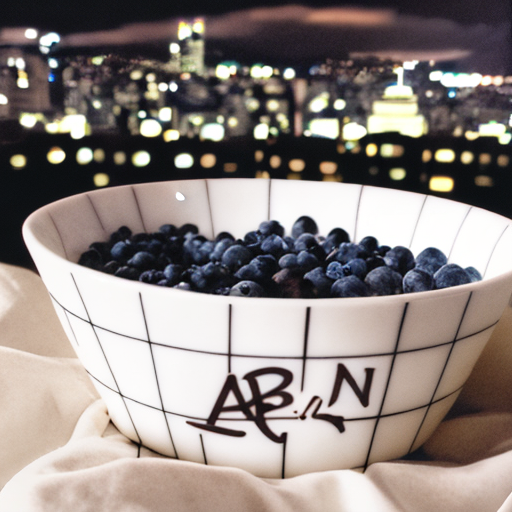}
    \includegraphics[width=0.13\textwidth]{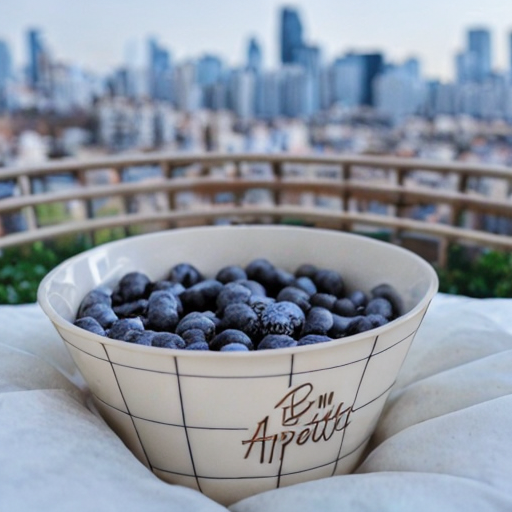}
    \\
    \makebox[\textwidth]{\sffamily "A V* bowl with the city in the background."}
    \\
    \includegraphics[width=0.13\textwidth]{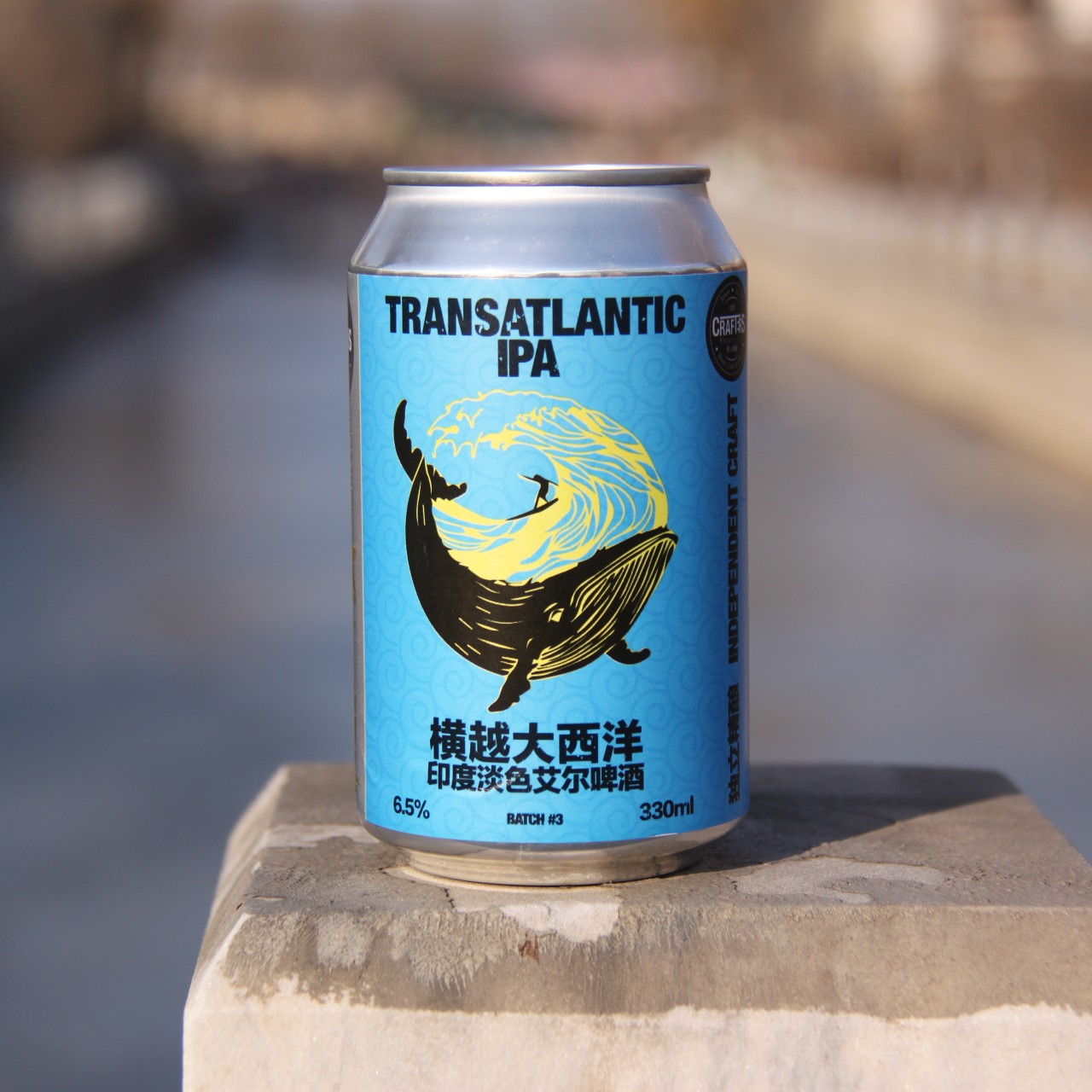}
    \includegraphics[width=0.13\textwidth]{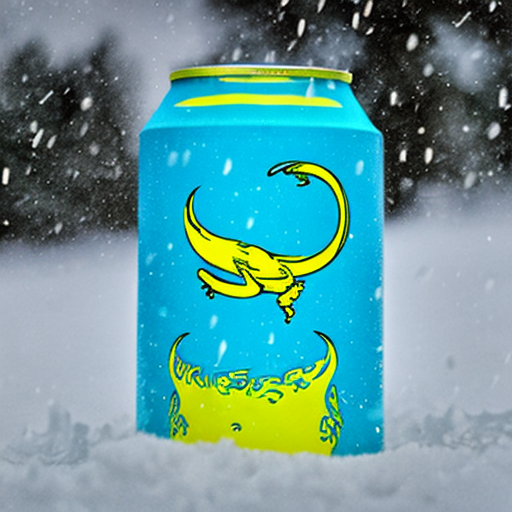}
    \includegraphics[width=0.13\textwidth]{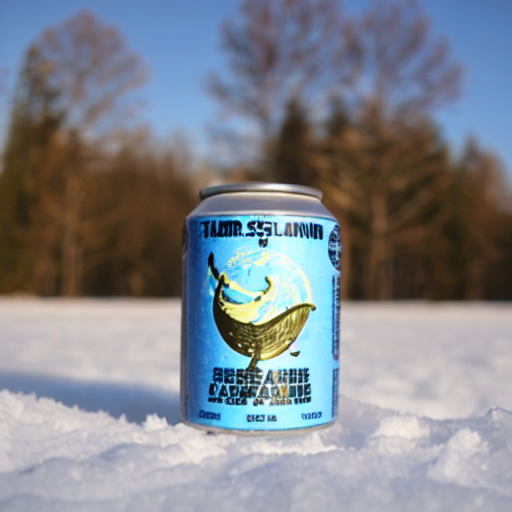}
    \includegraphics[width=0.13\textwidth]{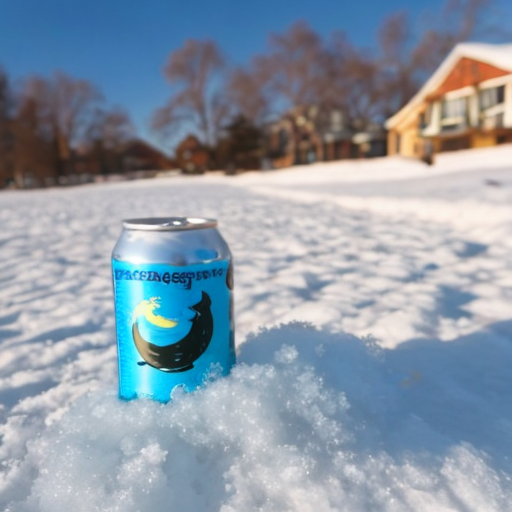}
    \includegraphics[width=0.13\textwidth]{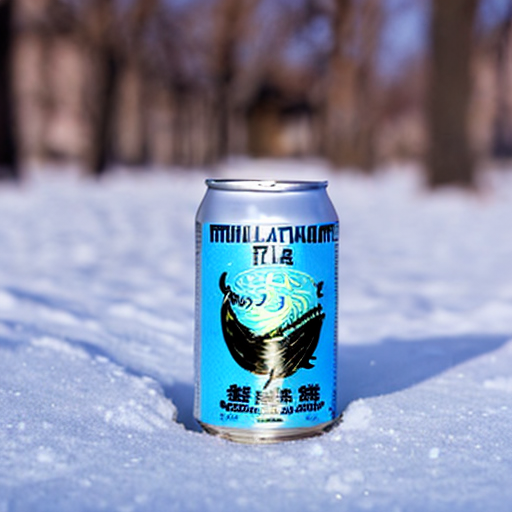}
    \includegraphics[width=0.13\textwidth]{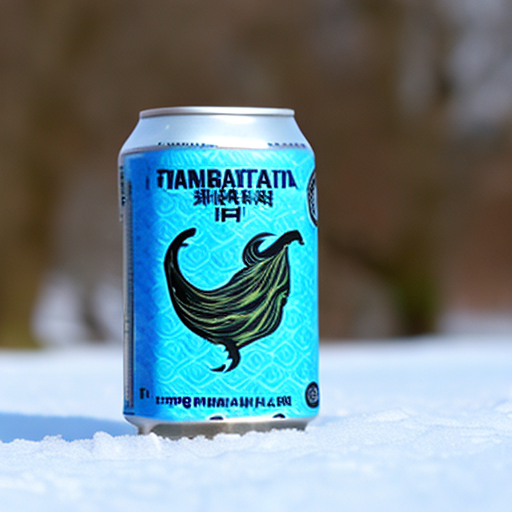}
    \includegraphics[width=0.13\textwidth]{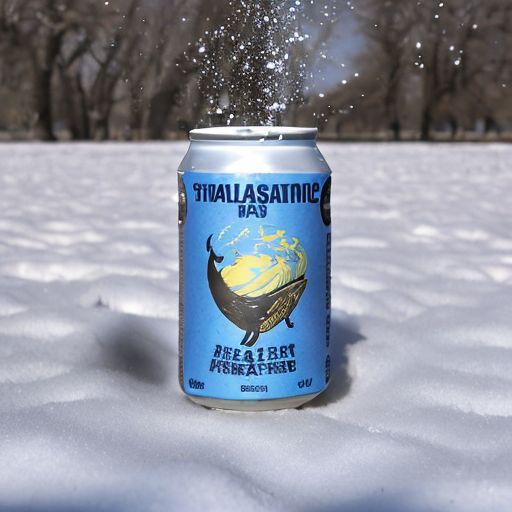}
    \\
    \makebox[\textwidth]{\sffamily "A V* can in the snow."}
    \\
    \includegraphics[width=0.13\textwidth]{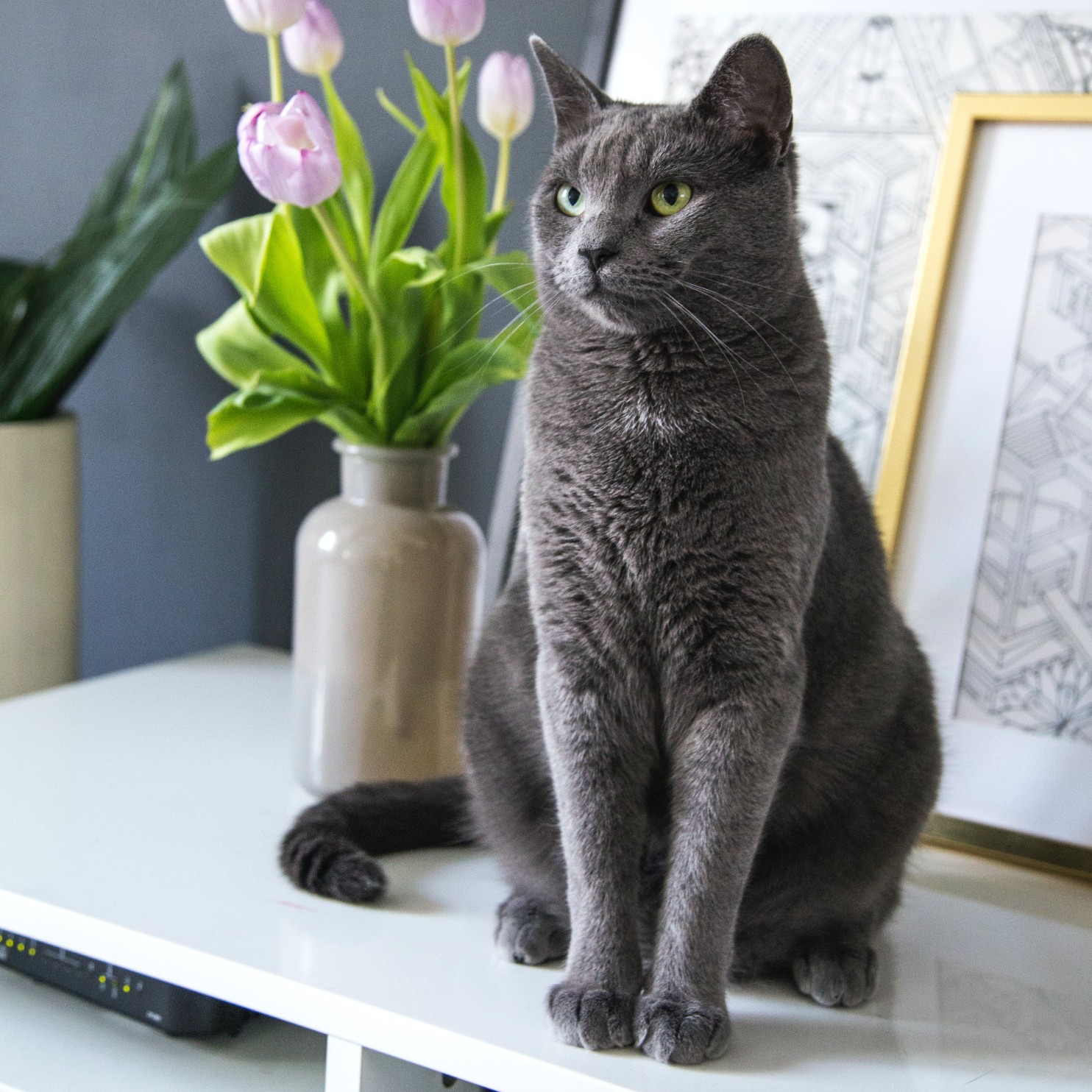}
    \includegraphics[width=0.13\textwidth]{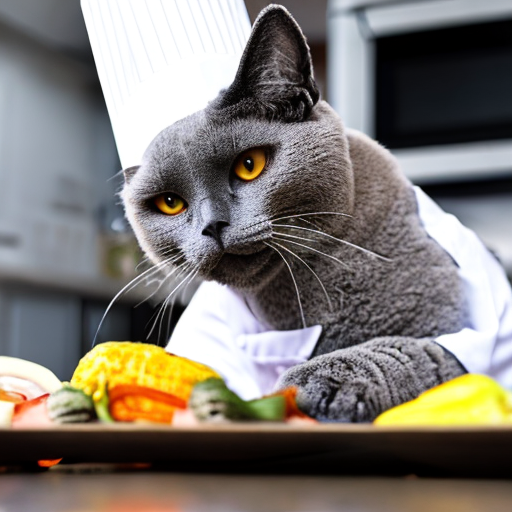}
    \includegraphics[width=0.13\textwidth]{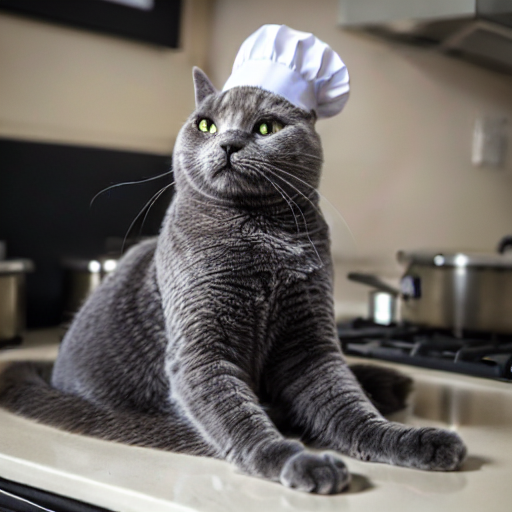}
    \includegraphics[width=0.13\textwidth]{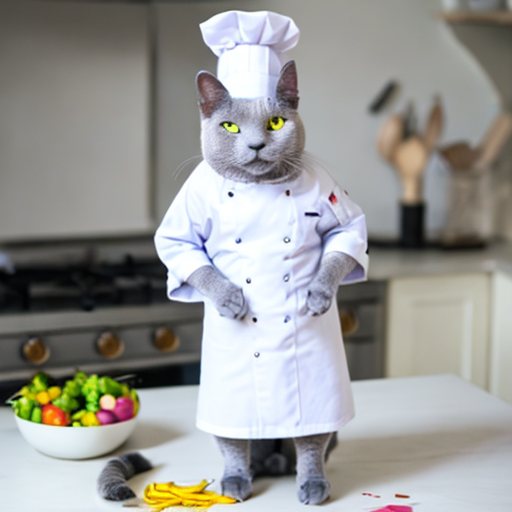}
    \includegraphics[width=0.13\textwidth]{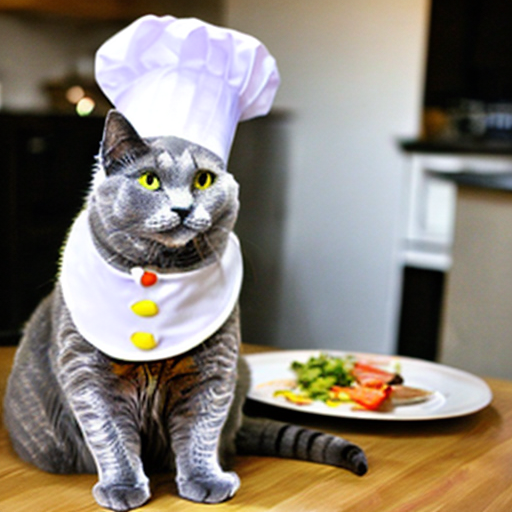}
    \includegraphics[width=0.13\textwidth]{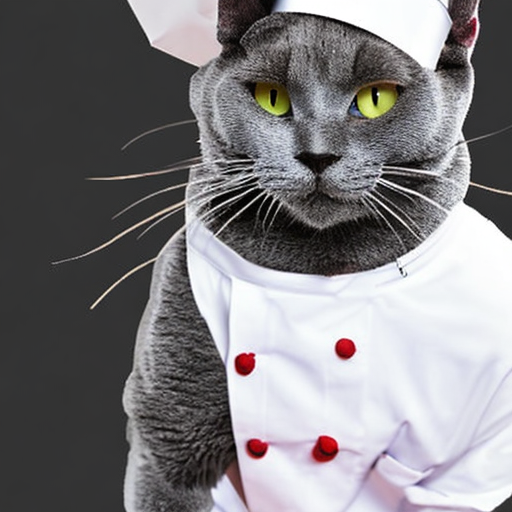}
    \includegraphics[width=0.13\textwidth]{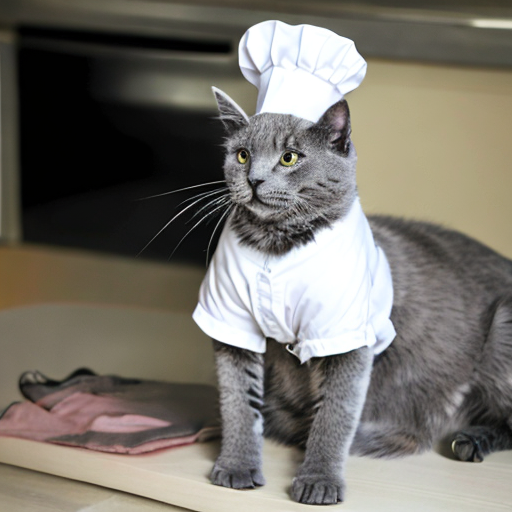}
    \\
    \makebox[\textwidth]{\sffamily "A V* cat in a chef outfit."}
    \\
    \includegraphics[width=0.13\textwidth]{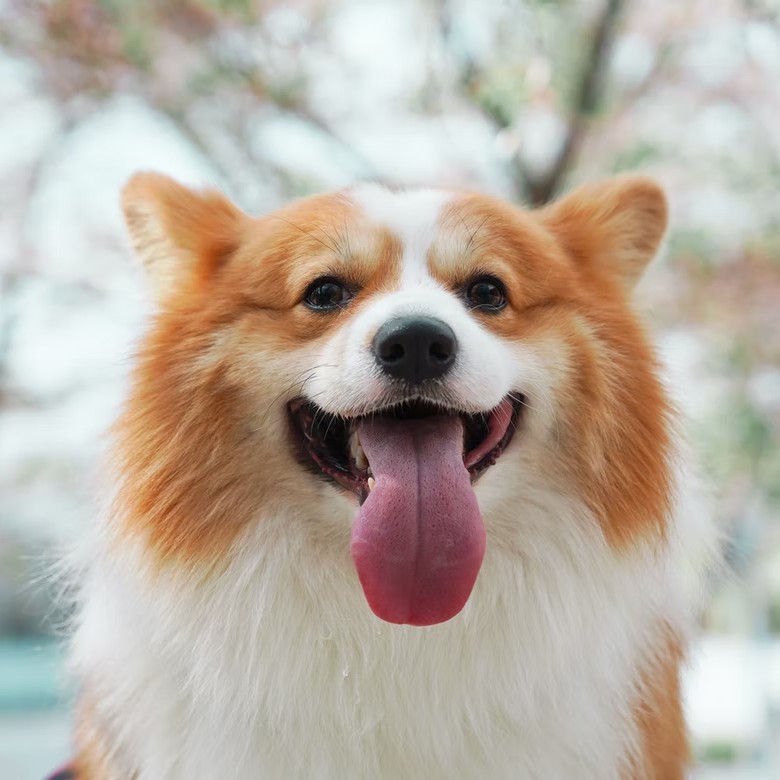}
    \includegraphics[width=0.13\textwidth]{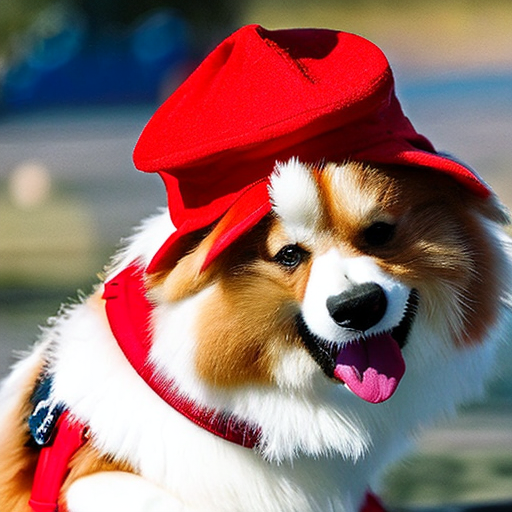}
    \includegraphics[width=0.13\textwidth]{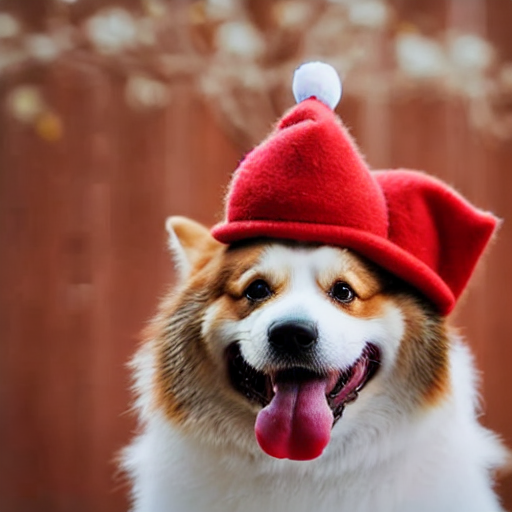}
    \includegraphics[width=0.13\textwidth]{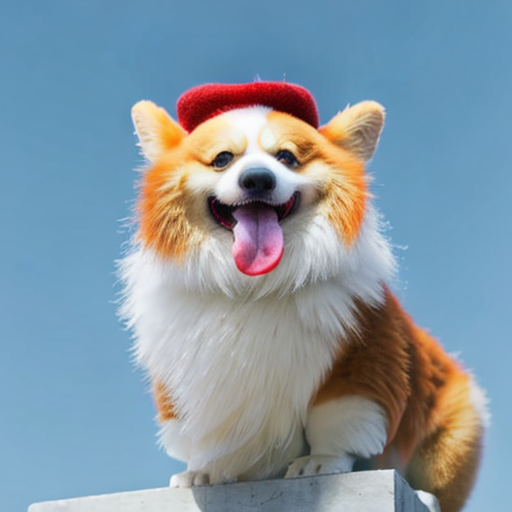}
    \includegraphics[width=0.13\textwidth]{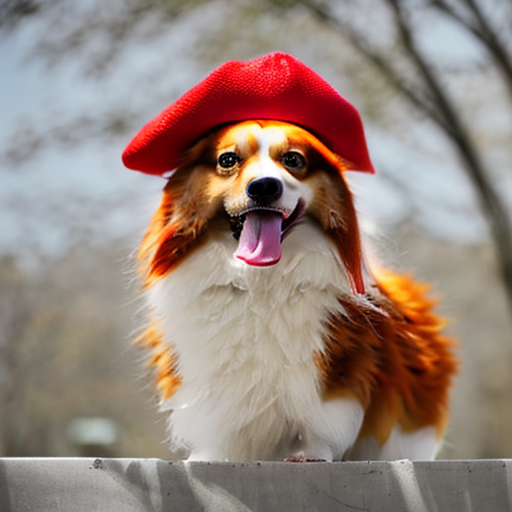}
    \includegraphics[width=0.13\textwidth]{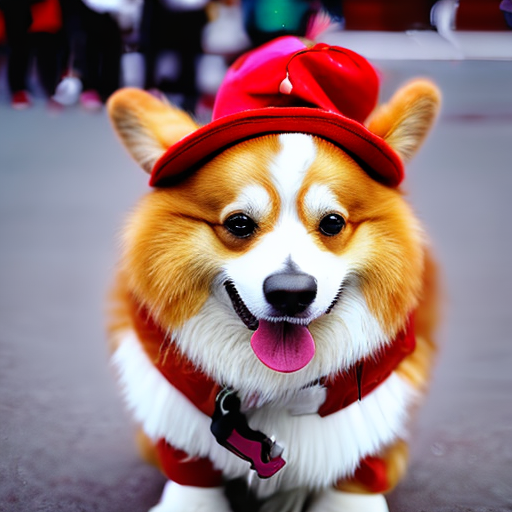}
    \includegraphics[width=0.13\textwidth]{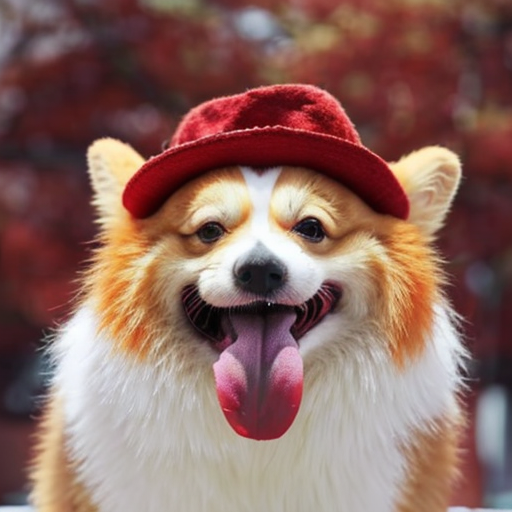}
    \\
    \makebox[\textwidth]{\sffamily "A V* dog wearing a red hat."}
    \\
    \includegraphics[width=0.13\textwidth]{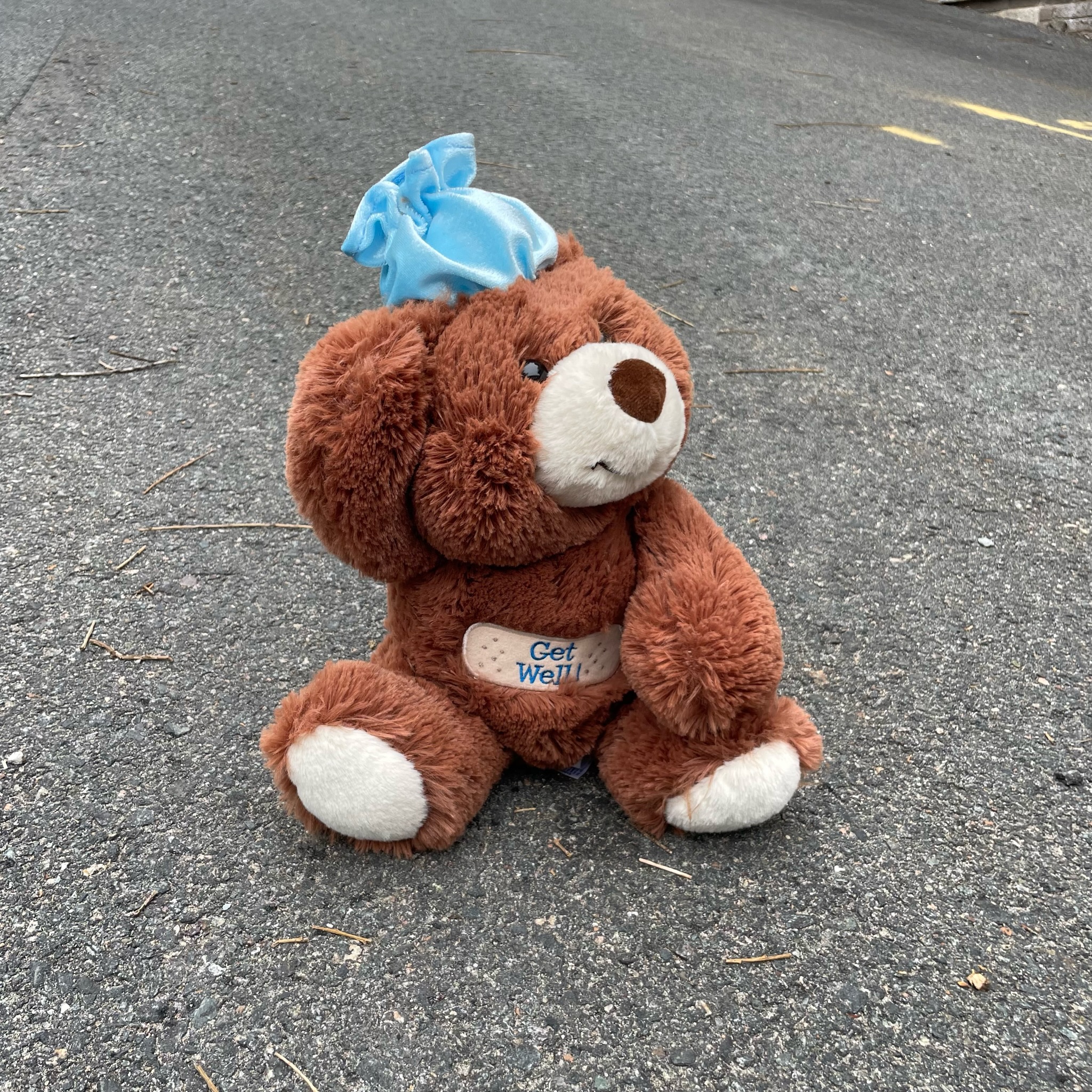}
    \includegraphics[width=0.13\textwidth]{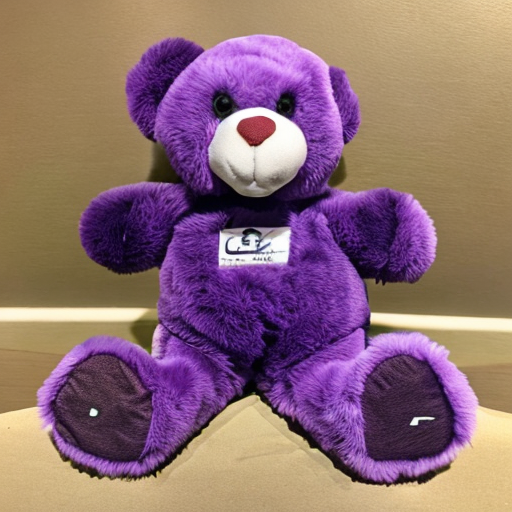}
    \includegraphics[width=0.13\textwidth]{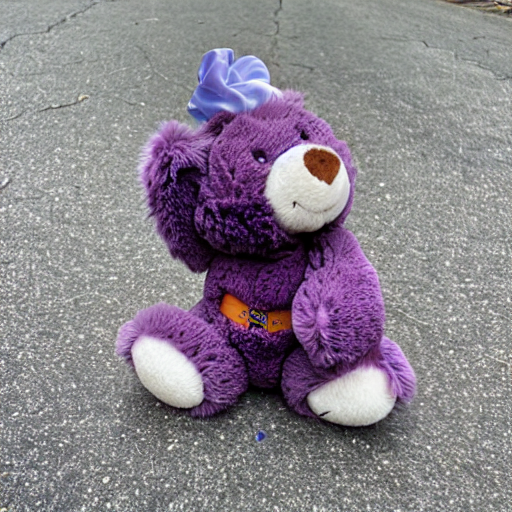}
    \includegraphics[width=0.13\textwidth]{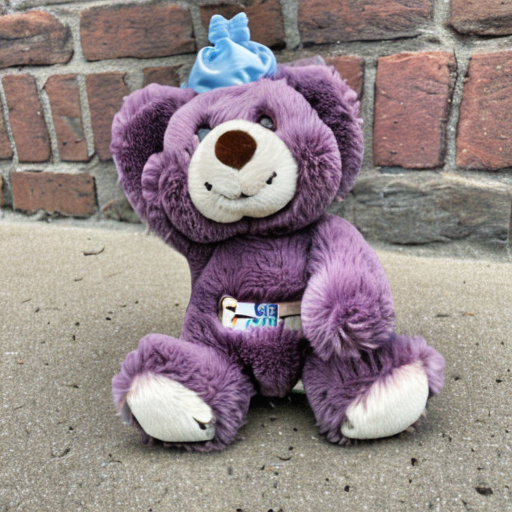}
    \includegraphics[width=0.13\textwidth]{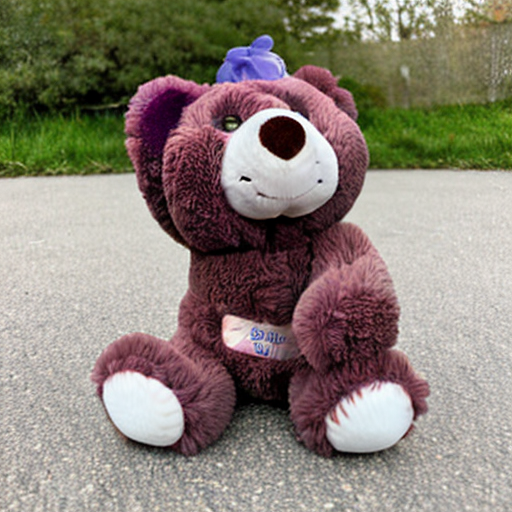}
    \includegraphics[width=0.13\textwidth]{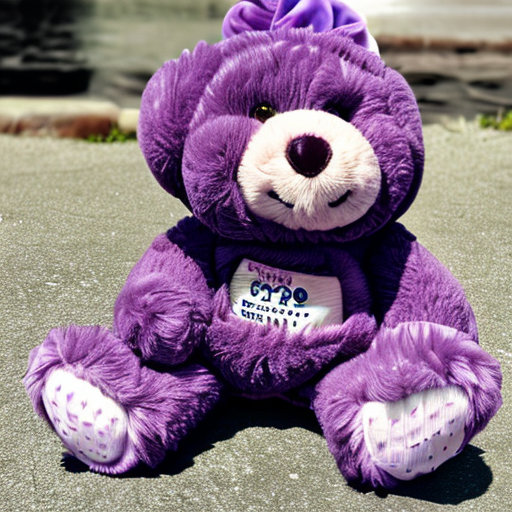}
    \includegraphics[width=0.13\textwidth]{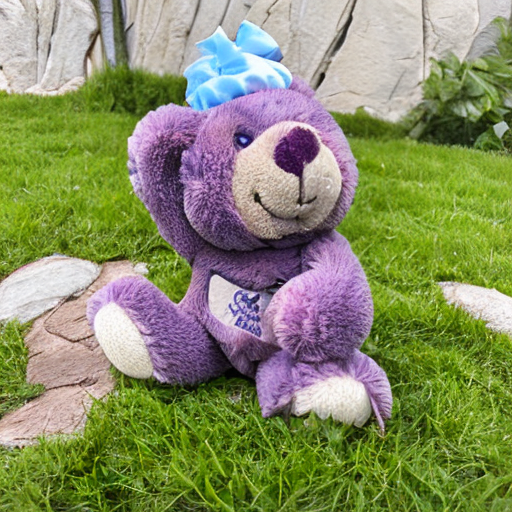}
    \\
    \makebox[\textwidth]{\sffamily "A purple V* stuffed animal."}
    \\
    \includegraphics[width=0.13\textwidth]{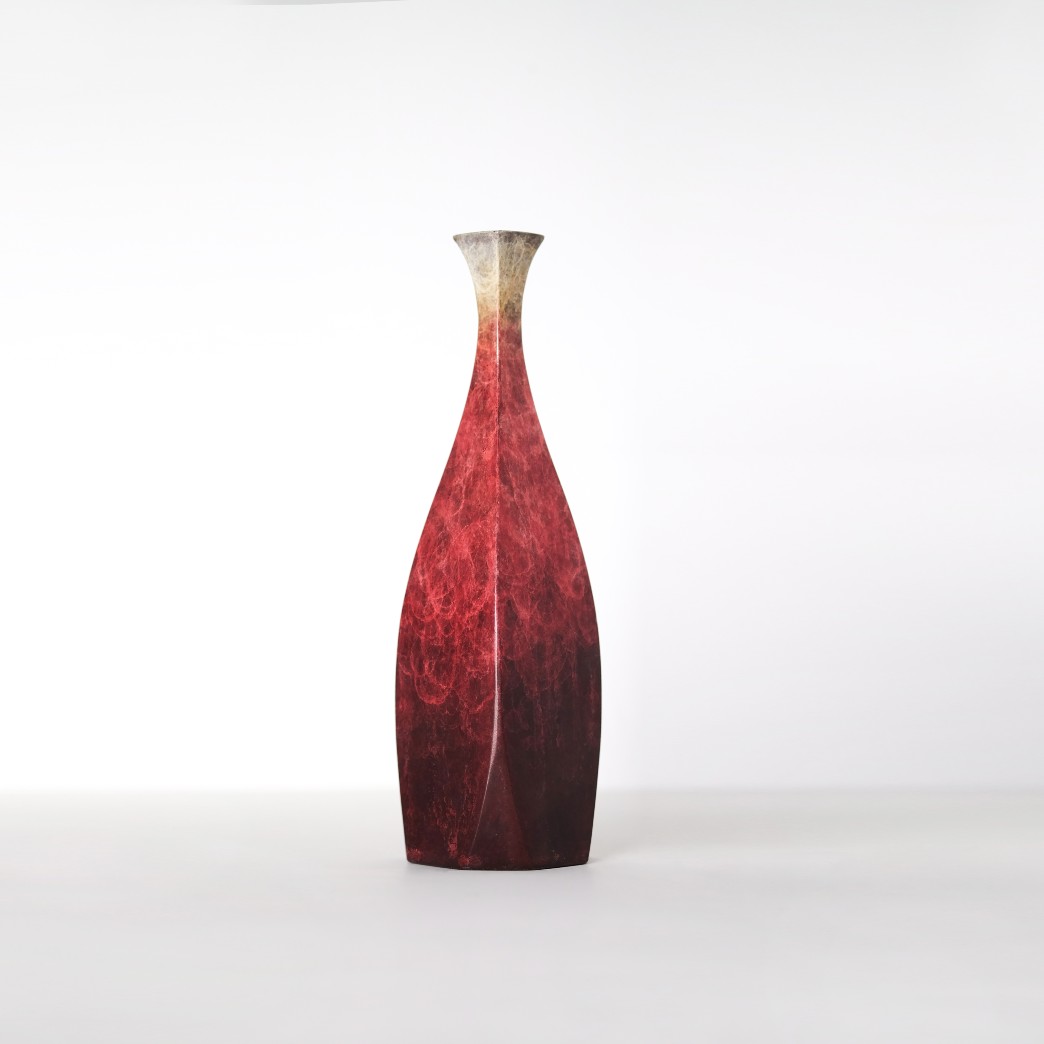}
    \includegraphics[width=0.13\textwidth]{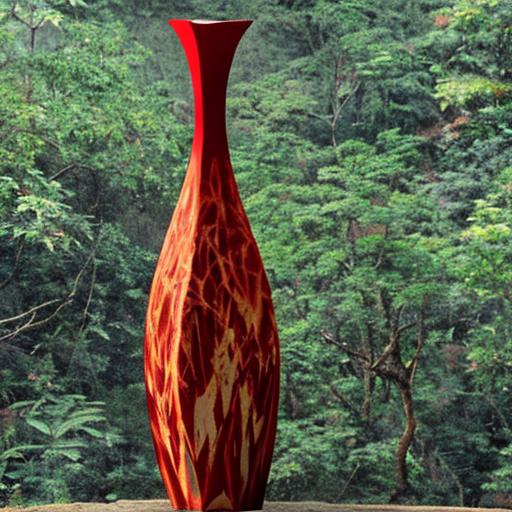}
    \includegraphics[width=0.13\textwidth]{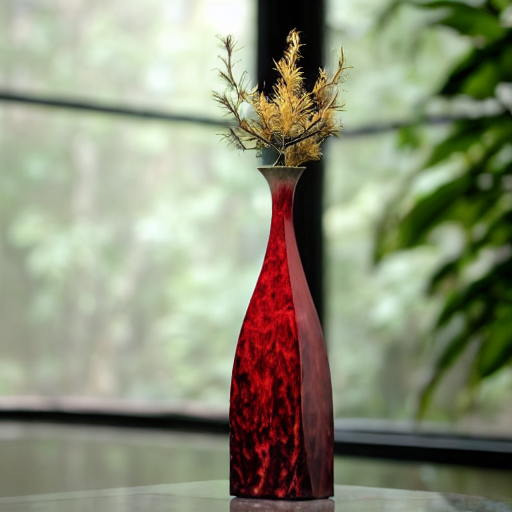}
    \includegraphics[width=0.13\textwidth]{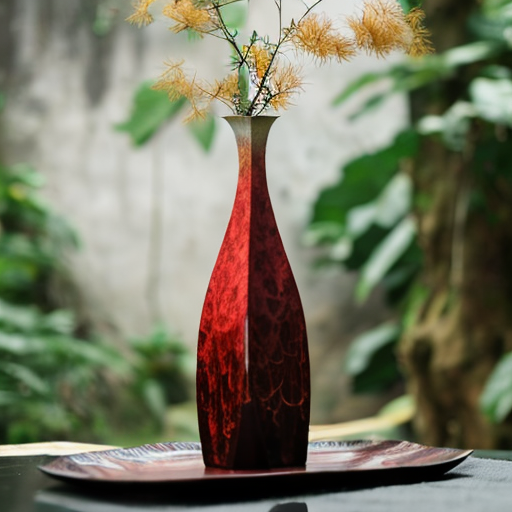}
    \includegraphics[width=0.13\textwidth]{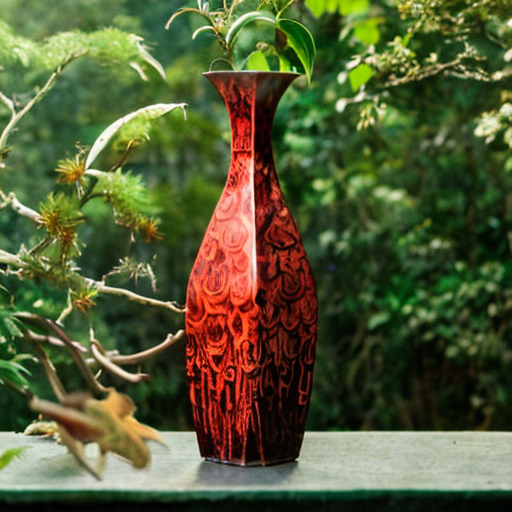}
    \includegraphics[width=0.13\textwidth]{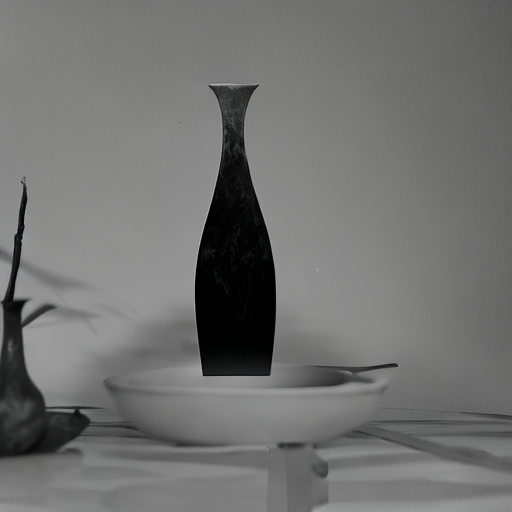}
    \includegraphics[width=0.13\textwidth]{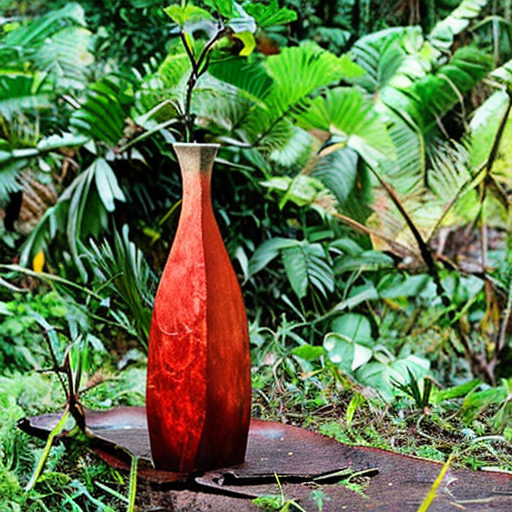}
    \\
    \makebox[\textwidth]{\sffamily "A V* vase in the jungle."}
    \\
    \caption{{\bf Qualitative result.} We compared our approach with current state-of-the-art methods, including Textual Inversion, DreamBooth, AttnDreamBooth, DisenBooth, and TextBoost, on the Dreambooth dataset. Our method demonstrates outstanding performance across multiple objects and animals, generating high-quality images with strong identity preservation and text alignment.} 
    \label{fig:qualitative_result4}
\end{figure*}
\subsection{Set up}

\noindent
{\bf Dataset.} We utilized the dataset proposed by DreamBooth~\cite{ruiz2023dreambooth}, which comprises $30$ distinct subjects. Each subject is associated with a collection of $4$ to $6$ images, and the dataset includes $25$ diverse text prompts that cover a variety of scenarios. Following the DreamBooth~\cite{ruiz2023dreambooth}, we generated $4$ images for each prompt associated with every subject, resulting in a total of $3,000$ images for comprehensive evaluation.

\noindent
{\bf Evaluation metrics.} We employed three primary metrics: CLIP-T, CLIP-I \cite{clip} and DINO \cite{dinov2}. 
CLIP-T measures the alignment of generated images with their text prompts by calculating the cosine similarity of their CLIP embeddings, with higher scores indicating a closer match between visual content and textual descriptions.
CLIP-I evaluates identity preservation by comparing the cosine similarity of CLIP embeddings between generated and real images, with higher scores suggesting greater similarity and effective identity retention.
DINO represents the average cosine similarity between the ViTS/16 DINO embeddings of generated and real images. A higher DINO score indicates greater similarity between the generated and input images.

\noindent
{\bf Baseline.} We compared our method with current state-of-the-art fine-tuning-based subject-driven text-to-image generation methods, including Textual Inversion \cite{gal2022textual_inversion}, Dreambooth \cite{ruiz2023dreambooth}, DisenBooth \cite{ruiz2023dreambooth}, AttnDreamBooth \cite{pang2024attndreambooth}, and TextBoost \cite{park2024textboost}.

\noindent
{\bf Implementation details.} Our implementation is based on the Stable Diffusion V$2.1$~\cite{rombach2022stable_diffusion}. During training, we utilize the AdamW optimizer with a learning rate set to $5 \times 10^{-4}$ and a batch size of $8$. The embeddings for $V^*$ are initialized using the corresponding embeddings for their respective categories, with the learning rate set to $1 \times 10^{-3}$ according to~\cite{pang2024attndreambooth}. The model trains for a total of 250 epochs. The values for $\lambda_1$, $\lambda_2$, and $\lambda_3$ are all set to $0.001$. The LoRA rank in both the U-Net and the text encoder is set to $4$. The number of experts is set to $2$. The experiments are conducted on a single NVIDIA A100 GPU.

\subsection{Comparison with Other Methods}

\noindent
{\bf Qualitative comparison.} To intuitively evaluate the performance of our proposed method, we conducted a qualitative comparison of images generated by various approaches. The selected subjects encompass common objects and animals, and we chose multiple representative text prompts that include background changes, appearance modifications, specified placements, and color alterations. The results are shown in Figure~\ref{fig:qualitative_result4}.

Our method outperforms existing approaches in terms of identity preservation while accurately capturing the scenes described in the text. We observe that Textual Inversion \cite{gal2022textual_inversion} exhibits weaker identity preservation due to its exclusive reliance on token embeddings. While DreamBooth \cite{ruiz2023dreambooth} generates high-quality outputs, it is prone to overfitting to specific scenes in the training dataset. AttnDreamBooth \cite{pang2024attndreambooth} achieves comparable results. Disenbooth \cite{chen2023disenbooth} performs poorly in terms of detail preservation. Additionally, TextBoost \cite{park2024textboost} is vulnerable to "augmentation leaking" because of its reliance on multiple data augmentation techniques. In contrast, our method effectively reproduces both the shapes and colors of objects, producing images that excel in detail and are robust to a variety of textual prompts. Whether the input involves simple descriptions or complex scenes, our approach consistently generates images that meet the specified requirements.

\begin{table*}[thb!p]
  \centering
  \begin{tabular}{ccccc}
    \toprule
    Method & CLIP-T($\uparrow$) & CLIP-I($\uparrow$) & DINO($\uparrow$) & storage($\downarrow$)\\
    \midrule
    Textual Inversion \cite{gal2022textual_inversion} & 0.257 & 0.733 & 0.473 & \textbf{7.5 kB} \\
    Dreambooth \cite{ruiz2023dreambooth} & 0.251 & 0.777 & 0.525 & 3.3 GB\\
    AttnDreamBooth \cite{pang2024attndreambooth} & \textbf{0.262} & 0.778 & 0.538 & 3.3 GB \\ 
    DisenBooth \cite{chen2023disenbooth} & 0.256 & 0.768 & 0.541 & 3.3 MB \\
    TextBoost \cite{park2024textboost} & 0.249 & 0.766 & 0.540 & 6.7 MB \\
    Ours & 0.260 & \textbf{0.789} & \textbf{0.546} & 9.8 MB \\
    \bottomrule
  \end{tabular}
  \caption{\textbf{Quantitative result.} Our method excels in text alignment, identity preservation, and detail retention, while maintaining a low storage requirement. Overall, our method provides an optimal trade-off between performance and efficiency.}
  \label{tab:quantitative_result}
\end{table*}

\noindent
{\bf Quantitative comparison.} We summarize the performance of our proposed method alongside several baseline approaches in Table~\ref{tab:quantitative_result}. 
Consistent with qualitative observations, Textual Inversion \cite{gal2022textual_inversion} performs well in text alignment but exhibits poor identity preservation. AttnDreamBooth \cite{pang2024attndreambooth} demonstrates solid performance across various metrics; however, it falls short in retaining identity details, resulting in a lower DINO score. Additionally, the final stage of AttnDreamBooth requires fine-tuning the entire UNet, consuming a substantial storage capacity of $3.3$ GB, similar to DreamBooth \cite{ruiz2023dreambooth}. DisenBooth \cite{chen2023disenbooth} and TextBoost \cite{park2024textboost} have lower storage requirements but lack effective identity preservation capabilities. In contrast, our method achieves an optimal balance between aligning textual descriptions and preserving the identity of the subjects, while only requiring a relatively small storage storage capacity of $9.8$ MB.

\subsection{Ablation Study}

\begin{figure}[thb!p]
\centering
\subfloat[w/o IEDM+FFM]{
            \label{fig:fig1.a}
		\includegraphics[width=0.21\textwidth]{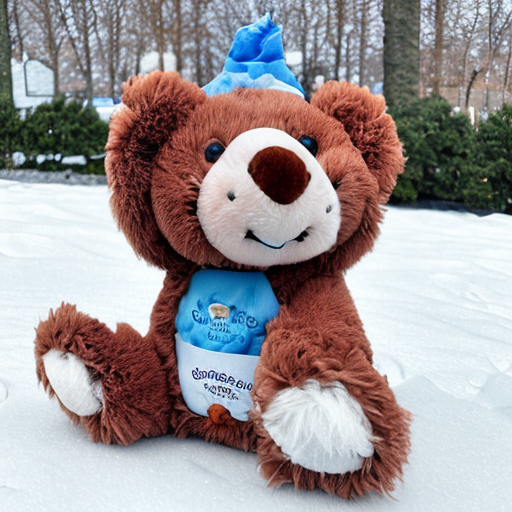}}
\subfloat[w/o FFM]{
            \label{fig:fig1.b}
		\includegraphics[width=0.21\textwidth]{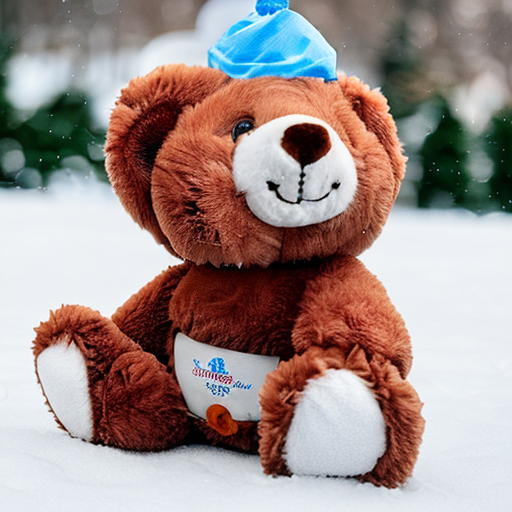}}
\\
\subfloat[w/o IEDM]{
            \label{fig:fig1.c}
		\includegraphics[width=0.21\textwidth]{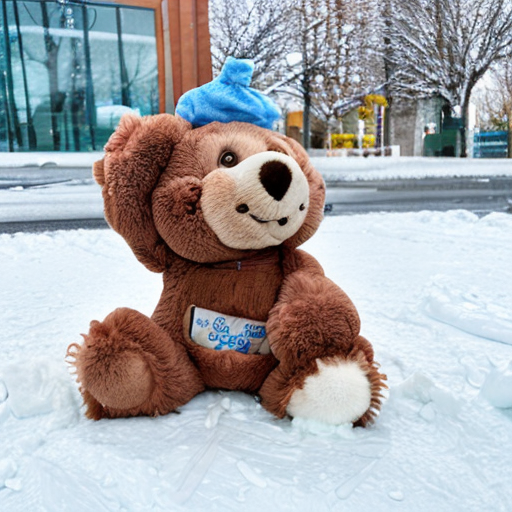}}
\subfloat[Full]{
            \label{fig:fig1.d}
		\includegraphics[width=0.21\textwidth]{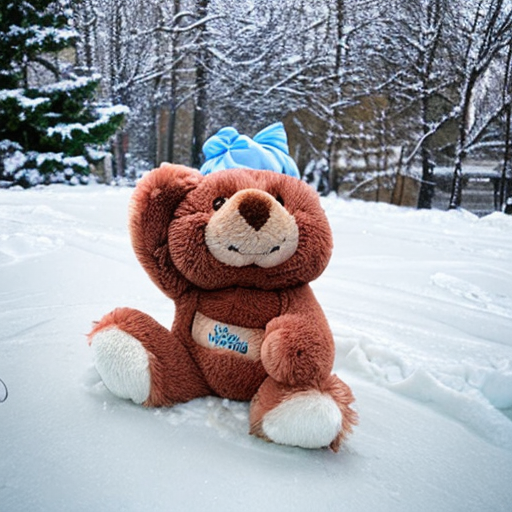}}
  
\caption{ {\bf Visualization of Ablation results.} We applied the prompt "a photo of a V* stuffed animal in the snow" to the specific subject "bear plushie.", illustrating the impact of different components of our proposed method.}
\label{fig:ablation_result}
\end{figure}

\noindent
\textbf{Abalation on ours proposed module.} To gain deeper insights into the contributions of various components in our proposed method, we conducted an ablation study focusing on the impact of the IEDM and the FFM. The results are summarized in Table~\ref{tab:ablation}.
Initially, we established a baseline model that excluded both IEDM and FFM. The results indicated significantly lower performance across all metrics, demonstrating that the absence of these two modules considerably weakens the overall capabilities of the model. 
Subsequently, we separately incorporated IEDM and FFM into the model. Each addition resulted in an improvement in performance; however, neither configuration surpassed the performance of the complete model that includes both modules.
Finally, when both IEDM and FFM were integrated, the model achieved its best performance, excelling across all evaluation metrics. These findings illustrate that both IEDM and FFM are crucial for enhancing model performance. IEDM effectively separates foreground and background features, while FFM enhances the adaptability of feature fusion, thereby ensuring the generation of high-fidelity images in diverse scenarios. We present the visual results in Figure~\ref{fig:ablation_result}.

\begin{table}
  \centering
  \begin{tabular}{cccc}
    \toprule
    Method & CLIP-T($\uparrow$) & CLIP-I($\uparrow$) & DINO($\uparrow$) \\
    \midrule
    w/o IEDM+FFM & 0.240 & 0.765 & 0.522 \\
    w/o IEDM & 0.253 & 0.777 & 0.529 \\
    w/o FFM & 0.245 & 0.779 & 0.541 \\
    our method & \textbf{0.260} & \textbf{0.789} & \textbf{0.546} \\
    \bottomrule
  \end{tabular}
  \caption{\textbf{Ablation study on the proposed IEDM and FFM modules.} Removing either module results in noticeable performance degradation, with the complete model performing best.}
  \label{tab:ablation}
\end{table}

\noindent
\textbf{Ablation on complementary loss.} We further evaluate the effectiveness of the three complementary loss terms $L_2, L_3, L_4$ by selectively removing each component. As shown in Table~\ref{tab:ablation_lambda}, the overall performance degrades progressively as more loss terms are excluded, indicating that these losses work collaboratively to enhance feature decoupling. Notably, removing $L_2$ leads to the largest performance drop, as this loss directly enforces the separation between identity-related and identity-irrelevant features. In contrast, models with only partial removal of the loss terms still retain competitive performance, suggesting that each loss contributes uniquely to the decoupling process. These results confirm that the proposed complementary losses are essential for effective feature disentanglement and contribute jointly to the final performance.

\begin{table}
  \centering
  \begin{tabular}{cccc}
    \toprule
    Method & CLIP-T($\uparrow$) & CLIP-I($\uparrow$) & DINO($\uparrow$) \\
    \midrule
    w/o $L_2+L_3+L_4$ & 0.245 & 0.768 & 0.524 \\
    w/o $L_2+L_3$ & 0.248 & 0.773 & 0.527 \\
    w/o $L_2+L_4$ & 0.249 & 0.775 & 0.529 \\
    w/o $L_3+L_4$ & 0.254 & 0.781 & 0.538 \\
    w/o $L_2$ & 0.251 & 0.777 & 0.531 \\
    w/o $L_3$ & 0.255 & 0.783 & 0.543 \\
    w/o $L_4$ & 0.257 & 0.785 & 0.544 \\
    our method & \textbf{0.260} & \textbf{0.789} & \textbf{0.546} \\
    \bottomrule
  \end{tabular}
  \caption{\textbf{Ablation study on the complementary loss functions.} Performance consistently improves as more losses are included, demonstrating the joint effectiveness of each loss in guiding feature disentanglement.}
  \label{tab:ablation_lambda}
\end{table}
\section{Conclusion}
\label{sec:Conclusion}
\indent
In this work, we introduced an innovative framework for subject-driven text-to-image generation that addresses critical challenges in disentangling identity-related and identity-irrelevant features while preserving alignment with textual descriptions. Our approach leverages a hybrid Implicit-Explicit foreground-background Decoupling Module (IEDM) and a Mixture of Experts-based Feature Fusion Module (FFM) to enhance feature separation and improve fusion adaptability. The IEDM achieves dual-level decoupling by combining feature-level implicit extraction of identity-irrelevant details with explicit separation of foreground and background using inpainting techniques, enabling more effective feature separation. The FFM then integrates background and foreground features, ensuring refined feature representation while mitigating potential interference from incomplete decoupling. Extensive experiments show that our method substantially enhances image generation quality, producing high-fidelity images that accurately capture textual descriptions and preserve subject identity across diverse scenes. Our work contributes a versatile solution for customizable text-to-image generation, advancing both quality and adaptability in personalized image synthesis tasks.

\newpage

{
    \small
    \bibliographystyle{ieeenat_fullname}
    \bibliography{main}
}


\end{document}